\documentclass[sigconf,screen,natbib=true]{acmart} 
\AtBeginDocument{%
  \providecommand\BibTeX{{%
    \normalfont B\kern-0.5em{\scshape i\kern-0.25em b}\kern-0.8em\TeX}}}

\copyrightyear{2023}
\acmYear{2023}
\setcopyright{rightsretained}
\acmConference[SOICT 2023]{The 12th International Symposium on Information and Communication Technology}{December 7--8, 2023}{Ho Chi Minh, Vietnam}
\acmBooktitle{The 12th International Symposium on Information and Communication Technology (SOICT 2023), December 7--8, 2023, Ho Chi Minh, Vietnam}\acmDOI{10.1145/3628797.3628976}
\acmISBN{979-8-4007-0891-6/23/12}

\RequirePackage{multirow}
\RequirePackage{subcaption}

\begin{document}

\title{Impact of Ground Truth Quality on Handwriting Recognition}

\author{Michael Jungo}
\authornote{Both authors contributed equally to this research.}
\email{michael.jungo@hefr.ch}
\affiliation{%
  \institution{University of Fribourg and HES-SO}
  \city{Fribourg}
  \country{Switzerland}
}

\author{Lars Vögtlin}
\authornotemark[1]
\email{lars.voegltin@unifr.ch}
\affiliation{%
  \institution{University of Fribourg}
  \city{Fribourg}
  \country{Switzerland}
}

\author{Atefeh Fakhari}
\email{atefeh.fakhari@unifr.ch}
\affiliation{%
  \institution{University of Fribourg}
  \city{Fribourg}
  \country{Switzerland}
}

\author{Nathan Wegmann}
\email{nathan.wegmann@unifr.ch}
\affiliation{%
  \institution{University of Fribourg}
  \city{Fribourg}
  \country{Switzerland}
}

\author{Rolf Ingold}
\email{rolf.ingold@unifr.ch}
\affiliation{%
  \institution{University of Fribourg}
  \city{Fribourg}
  \country{Switzerland}
}

\author{Andreas Fischer}
\email{andreas.fischer@hefr.ch}
\affiliation{%
  \institution{University of Fribourg and HES-SO}
  \city{Fribourg}
  \country{Switzerland}
}

\author{Anna Scius-Bertrand}
\email{anna.scius-bertrand@hefr.ch}
\affiliation{%
  \institution{University of Fribourg and HES-SO}
  \city{Fribourg}
  \country{Switzerland}
}

\renewcommand{\shortauthors}{Jungo and Vögtlin, et al.}

\begin{abstract}
  Handwriting recognition is a key technology for accessing the content of old manuscripts, helping to preserve cultural heritage. Deep learning shows an impressive performance in solving this task. However, to achieve its full potential, it requires a large amount of labeled data, which is difficult to obtain for ancient languages and scripts. Often, a trade-off has to be made between ground truth quantity and quality, as is the case for the recently introduced Bullinger database. It contains an impressive amount of over a hundred thousand labeled text line images of mostly premodern German and Latin texts that were obtained by automatically aligning existing page-level transcriptions with text line images. However, the alignment process introduces systematic errors, such as wrongly hyphenated words. In this paper, we investigate the impact of such errors on training and evaluation and suggest means to detect and correct typical alignment errors.
\end{abstract}



\begin{teaserfigure}
  \includegraphics[width=\textwidth]{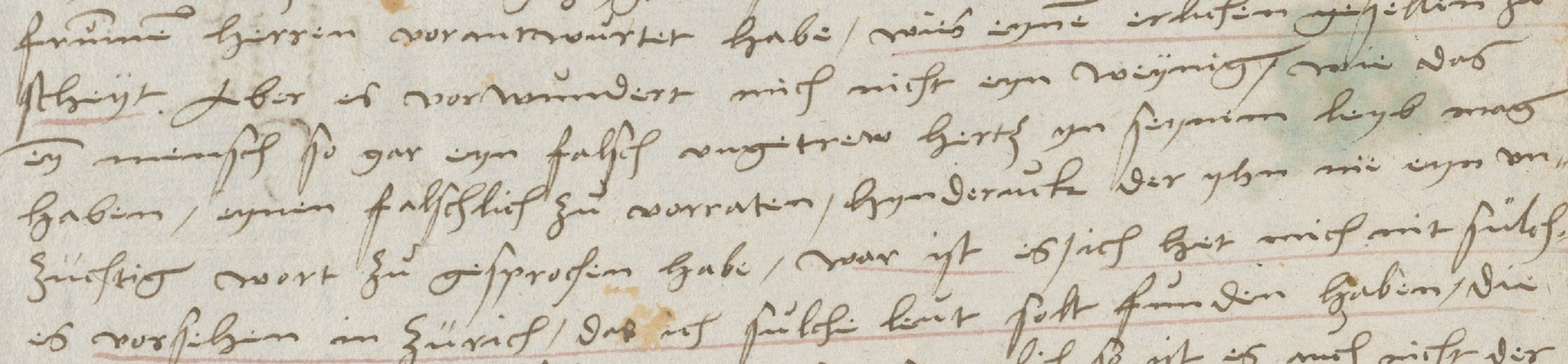}
  \caption{An extract of a letter from the Bullinger dataset.}
  \label{fig:bullinger_db_page}
\end{teaserfigure}

\maketitle

\section{Introduction}

Libraries worldwide are in the process of digitizing millions of handwritten documents to preserve our cultural heritage. In order to access the textual content of these documents and make them available in digital libraries, handwritten text recognition (HTR) is needed. The automatic transcription of handwriting images to machine-readable text is a challenging problem, considering the large number of historical scripts, languages, writing supports, and writing instruments used~\cite{fischer2020handwritten}. Furthermore, many ancient languages are only known to experts, which makes it difficult to obtain learning samples for training machine learning algorithms.

Similar to other applications of machine learning and pattern recognition, HTR technology has evolved from early rule-based algorithms over machine learning-based methods, such as hidden Markov models~\cite{fink2007use}, towards deep learning-based approaches. The latter omit manual feature engineering and allow end-to-end learning from the raw images to machine-readable transcriptions. Typical deep neural network architectures for HTR are based on convolutional backbones to extract features and recurrent layers for sequential recognition of characters and words~\cite{laia}. More recently, Transformer architectures have demonstrated state-of-the-art HTR performance~\cite{trocr}.

However, end-to-end learning using deep neural networks has a significant drawback, which is the large amount of training data needed. For HTR, learning samples typically consist of text line images together with their transcriptions. With a view to the high effort of creating such ground truth samples with the help of experts, a typical trade-off has to be made between data quality and data quantity.

An example of such a trade-off is the recently introduced Bullinger handwriting dataset~\cite{scius2023bullinger}. It is one of the currently largest HTR research datasets and contains thousands of letters written and received by Heinrich Bullinger, an important Swiss reformer of the 16th century. The database includes over one hundred thousand text line images together with their transcriptions. However, in order to obtain such a high data quantity, an automatic alignment between existing page-level transcriptions and automatically extracted text line images was performed, at the expense of data quality. Multiple errors are introduced in the ground truth, including wrong hyphenations, i.e.,\ parts of hyphenated words that are on the wrong text line, abbreviations, that are written out in full and therefore differ from the image, wrong capitalization, and wrong punctuation marks. When using the Bullinger dataset for HTR research, both training and evaluation are affected by these inconsistencies in the ground truth.

In related fields, several studies have been conducted to investigate the impact of ground truth quality on deep learning, for example in the context of object detection~\cite{agnew2023quantifying,li2022testing}, text-line segmentation~\cite{alberti2019labeling,scius2019layout}, and semantic segmentation~\cite{rahal2023historical,taran2020impact} in natural images or historical document images. However, the problems encountered for HTR are specific and to the best of our knowledge, there are currently no comprehensive studies on the impact of ground-truth quality for deep learning-based HTR.

In this paper, we specifically investigate the impact of alignment errors on the HTR performance in the context of the Bullinger handwriting dataset. We propose several methods for detecting typical alignment errors and provide a new test set for the Bullinger dataset that has been manually corrected and is publicly available~\footnote{\url{https://tc11.cvc.uab.es/datasets/BullingerDB_1}}. In several experiments, we demonstrate the effect of errors in the ground truth on state-of-the-art HTR models.

The remainder is organized as follows: in \autoref{sec:database} we present the Bullinger database, in \autoref{sec:method} we describe the methodology used to filter the training set and correct the test set, in \autoref{sec:experimentation} the experimentation with the results are shown, then we conclude in the last section.

\section{Bullinger Database}
\label{sec:database}
Heinrich Bullinger (1504\textemdash{}1575) was an important Swiss reformer, who had a comprehensive letter correspondence with more than 1\,000 people in Europe. 
The Bullinger Digital project aims to collect images of existing letters and transcriptions in a single database and make them publicly accessible. 
Bullinger's correspondence is composed of around 10\,000 letters he received and around 2\,000 letters written by himself. The letters are mainly written in Latin followed by premodern German, and some parts in Greek, Italian, French, and Hebrew. 

Multiple sources of transcriptions at the page level were collected. The alignment was performed automatically using the Text2Image module of the Transkribus~\cite{transkribus}. An example of an extract of a page is shown in \autoref{fig:bullinger_db_page}. This paper considers the recently published database for handwriting recognition research~\cite{scius2023bullinger}, which contains 3\,622 letters with more than 150\,000 text lines. More detailed statistics on the database and the standard splits into training, validation, and testing for HTR research are provided in \autoref{tab:data_split}. 

\begin{table}[h]
\centering
\begin{tabular}{lrrrr}
\toprule
\textbf{}  & \textbf{Training} & \textbf{Validation} & \textbf{Test} & \textbf{Total} \\
\midrule
\# of letters   & 2\,581 & 337  & 704 & 3\,622 \\ 
\# of pages  & 5\,927  & 806  & 1\,660 &  8\,393 \\ 
\# of lines   & 109\,627  & 14\,516 & 31\,103 & 155\,246 \\ 
\# of words   & 876\,003  & 122\,211 & 243\,500 & 1\,241\,714 \\
\bottomrule
\end{tabular}
\vspace{1em}
\caption{\textbf{Database Statistics.} Distribution of writers, letters, pages, text lines, and words for the different splits.}
\label{tab:data_split}
\end{table}

\begin{figure*}
    \centering
    \begin{subfigure}[b]{0.23\textwidth}
        \centering
        \includegraphics[height=4em]{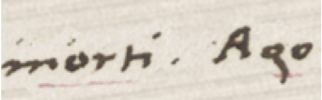}
        \caption{\\Ago \\ morti. Ago}
        \label{fig:errors_gt_a}
    \end{subfigure}
    \hspace{2em}
    \begin{subfigure}[b]{0.26\textwidth}
        \centering
        \includegraphics[height=4em]{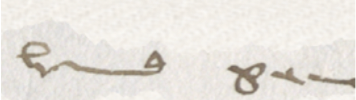}
        \caption{\\hand genommen \\ hand genen}
        \label{fig:errors_gt_b}
    \end{subfigure}
    \hspace{0.2em}
    \begin{subfigure}[b]{0.18\textwidth}
        \centering
        \includegraphics[height=4em]{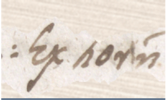}
        \caption{\\; ex horum \\ .Exu}
        \label{fig:errors_gt_c}
    \end{subfigure}
    \hspace{-3.3em}
    \begin{subfigure}[b]{0.3\textwidth}
        \centering
        \includegraphics[height=4em]{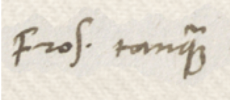}
        \caption{\\Froschouerus tanquam\\Froscu tanqum}
        \label{fig:errors_gt_d}
    \end{subfigure}
    \caption{\textbf{Example of ground truth errors.} From top to bottom: excerpt from the image of the text line, ground truth, and prediction by a fully trained HTR model.}
\end{figure*}

Due to automatic alignment and human transcription errors, which cannot be corrected by human intervention, distortions appear between the ground truth and the text on the image. One frequent source of errors is hyphenated words, where one part of a word is written at the end of a line and the other part begins on the next line. In the ground truth, the complete word not may appear on the first or second line, as shown in \autoref{fig:errors_gt_a} and \autoref{fig:errors_gt_b}. Another source of error is related to punctuation and capitalization. An example is shown in \autoref{fig:errors_gt_c}. Finally, there are abbreviations to consider, as shown in \autoref{fig:errors_gt_d}. Some abbreviations are common across multiple authors, while others are invented \textit{ad hoc}, for example, to abbreviate names.

\section{Methods} 
\label{sec:method}

For the purpose of detecting typical alignment errors, we suggest two methods for identifying hyphenated words and one method for distinguishing outliers among the labeled text lines in general. 
Finally, we describe the handwriting recognition systems that will be used to experimentally evaluate the impact of ground truth errors. 

\newpage
\subsection{Hyphenated Words}
\label{sec:hyphenated-words}

When there is not enough space at the end of the line to write the last word completely, it has often been
written in part before being continued on the following line by hyphenating the word. Usually, a hyphen should be added at
the end of the line where the word started. In the letters written during this period, an equal sign (=) was
used instead of the dash (\textendash{}) that is used nowadays. Even though the equal sign is expected, it can vary
widely in appearance, where it is sometimes written in a slanted manner, making it more difficult to be identified as an
equal sign. Furthermore, due to degradation of the physical letter or other quality issues, it may either be only
partially visible, to the point where it looks like a colon (:), or not be visible at all. Therefore, it is
necessary not to rely on a hyphen symbol being present.

The alignment of the Bullinger dataset was originally done with Transkribus, which led to
certain behavioral characteristics when it comes to aligning the transcription with the text lines. Most notably, the
hyphenated words are assigned completely to one line and the leftover fragments on the other line are entirely removed.
While it tends to be the case, that the full word is put at the end of the line, where the word started, and the suffix
on the following line is ignored, there are also occurrences of the inverse, where the word is put fully on the next
line. To make matters more complex, sometimes the prefix (or suffix) remains intact, which is mostly due to the fact, that it is
also a full word on itself, whereas the other part is then also completed to become a full word.

In order to detect candidates of lines containing hyphenated words, we have considered the following two approaches:
Firstly, comparing the HTR result with the ground truth to see where there is an addition or deletion at either the end
of the line or the beginning of the line, and secondly, using a dedicated hyphenation detection model.

\subsubsection{HTR Based Hyphenation Detection}

Since the HTR models contain an internal language model, or in the case of TrOCR, it is technically a fully pre-trained
language model used as a decoder, they often try to complete a word whenever possible, especially when trained on data
that has essentially no fragments of words in the ground truth. By exploiting this property, we can find words that have
been erroneously left out of the ground truth, as the HTR model, which is applied to a single line, will not have
the context of the other line to know whether or not that part of the word has already been transcribed.

To detect the potential hyphenation candidates, the transcription of the HTR model is compared with the ground truth.
A line is considered a candidate when there is either an addition/deletion of at least three characters at the end of
the line or an addition/deletion of at least three characters at the start of the following line. It is important to
note, that an addition/deletion means that there is no substitution and merely more characters are in either the
transcription or the ground truth, i.e.\ a prefix or suffix has been added. For this purpose, it does not matter whether
the transcription of the HTR model or the ground truth contains more characters. To avoid false positives, such as
having a comma at the end or misinterpreting an undesired line as an additional character, there must be at least at
difference of \textit{three characters}, in which case it is guaranteed to be a real prefix or suffix.

\subsubsection{Hyphenation Detection Model}

For the second approach, a dedicated model is trained to find words that are hyphenated at the end of the line as proposed by~\cite{andres2023search}. In order
to train such a model, a dataset is required that already contains the information about hyphenated words. The
\textit{Finnish Court Records Dataset}~\cite{finnish-dataset} is part of the Finnish Court Records collection held by
National Archives of Finland and contains 40k images of Finnish text lines with their transcriptions. Around
a quarter of the dataset consists of lines where the last word is hyphenated to be finished on the next line. Whether or
not there is an actual hyphen (\textendash{}) present in the image, the dataset always indicates a hyphenated word at
the end of the line with the symbol ¬. That means that the models need to learn the concept of hyphenated words
rather than relying on detecting a hyphen symbol, which is particularly interesting for historical documents since
other symbols may have been used at the time or they are simply not present, whether by choice or due to the
deterioration of the physical document.

\subsection{Deviation between Ground Truth and Prediction}

Errors in the content of the ground truth can lead to a significant discrepancy between the number of characters in a prediction made by a model, trained to recognize the content of a line, and the ground truth. For example, if on the line image, there is the beginning of a hyphenated word but the ground truth contains the whole word, or conversely, if the network recognizes part of a word on the image when it is not part of the ground truth, then there will be a discrepancy between the number of character tokens predicted and the ground truth. Another source of error comes from abbreviations that are not commonly accepted: for example, the image shows the first letter of a first name, whereas the ground truth shows the whole first name. This problem is not significant for common abbreviations, as the network is capable of learning them. One solution to automatically exclude lines containing such errors from the training set is to exclude lines with the greatest discrepancy between prediction and ground truth. To do this, we compare the predictions of a previously trained network and retain the lines with the number of characters closest to that of the ground truth.

\subsection{Annotation Platform}

Annotating any data by hand is very time-consuming, therefore it is essential to have a labeling tool or platform to
make the annotation process as efficient as possible. We opted for Label Studio~\cite{label-studio}, an open-source data
labeling platform, that can be deployed on a Kubernetes cluster and accessed by all annotators through a configurable
web interface.

The primary objective of this annotation effort was the correction of lines containing errors related to hyphenated
words at the end of the line. As briefly mentioned in \autoref{sec:hyphenated-words}, the alignment of the transcription
with the image lines caused issues for the hyphenated words, since the word is not split up into the two lines, but
rather assigned completely to one line and on the other line the remaining part of the word is completely ignored. This
results in some lines having more characters, or even words, than their image actually contains, while others have
fewer than expected.

In the labeling interface, the image of the text line is displayed alongside the ground truth, which can be edited by
the annotator to fix the errors. Additionally, there are labels that should be selected to indicate the issues
identified in that line. Firstly, the status of the annotations must be selected from one of the four labels:
\textit{Correct}, \textit{Fixed}, \textit{Unsure} and \textit{Has Errors}. Secondly, the changes in regard to the
hyphenated words need to be categorized with the following options, for the start and end of the line, respectively:

\begin{itemize}
  \item \textbf{Missing Word(s):} Any full word had to be \textit{added} to the transcription at the start/end of the line.
  \item \textbf{Additional Word(s):} Any full word had to be \textit{removed} from the transcription at the start/end of the line.
  \item \textbf{Hyphenated (missing):} Part of the hyphenated word had to be \textit{added} to the transcription at the start/end of the line, i.e., \ the hyphenated part was missing from this line.
  \item \textbf{Hyphenated (extra chars):} Part of the first/last word had to be \textit{removed} from the transcription at the start/end of the line, i.e., \ the hyphenated word was erroneously completed.
  \item \textbf{Hyphenation Character}\footnote{This only applies to the \textit{end of line}} There is a hyphenation character, dash (\textendash{}) or equal (=) or whichever symbol is used as a hyphen, at the end of the line in the image.
\end{itemize}
The aforementioned options must be selected separately for the start and end of the line since it is possible to have
the hyphenation from the previous line and a new one at the end of this one. In situations where it might be unclear
whether it is a suffix of a hyphenated or full word on its own, the annotation page contains a direct link to
the full letter on Bullinger Digital~\cite{bullinger-digital}, the online archive of all the letters, which can be used
as a cross-reference.

\subsection{Handwriting Recognition}

Two fundamentally different HTR models that are both considered state of the art in HTR, are used for handwriting recognition. 
PyLaia~\cite{laia} is a more traditional CRNN, with a stack of convolutions, similar to VGG~\cite{vgg}, followed by a bidirectional LSTM~\cite{lstm} and a Connectionist Temporal Classification (CTC)~\cite{ctc} for the decoding part.
TrOCR decided to combine a pre-trained Vision Transformer (ViT)~\cite{vit} with a pre-trained language model, such as BERT~\cite{bert}, and since they are sharing the Transformer architecture, they can easily be combined into one model. 
For the encoder, BEiT~\cite{beit} is used, which was pre-trained with the masked image modeling task, and the decoder is a pre-trained RoBERTa~\cite{roberta}.
By combining two Transformer models, TrOCR is significantly larger than PyLaia in terms of parameters.

\section{Experimentation}
\label{sec:experimentation}

\subsection{Setup}

All experiments were conducted with models implemented in PyTorch~\cite{pytorch}. We have used TrOCR from
HuggingFace's Transformers library~\cite{huggingface} with its default configuration and the cross-entropy as the loss
function and PyLaia. For both, AdamW~\cite{adamw} is used as an optimizer with a weight decay of $10^{-4}$, $\beta_1 = 0.9$ and $\beta_2
= 0.98$. 
For TrOCR, the learning rate is warmed up over 500 steps by increasing it linearly to reach a peak learning rate of $2
\cdot 10^{-5}$. Afterwards, it follows the learning rate schedule proposed in~\cite{transformer}, which decays the
learning rate by the inverse square root of the number of iterations. As is common practice, an Exponential Moving
Average (EMA)~\cite{ema} is applied during the training to obtain the final weights of the model.
For PyLaia, we keep the learning learning rate constant at $5.5\cdot 10^{-4}$. To ensure comparability with the baseline proposed here~\cite{scius2023bullinger}, we have kept the same parameters as described in the paper. 

To evaluate the HTR performance, a standard measure of character error rate (CER) and word error rate (WER) is used. They are calculated by computing the string edit distance between the recognition output and the ground truth, to obtain the number of substitution, deletion, and insertion errors. We obtain the CER and WER by dividing the number of character/word errors by the number of characters/words in the ground truth.

\subsection{Filtering the Training Set}
With the aim to improve the quality of the training set, we applied the previously introduced methods to filter out samples with a low-quality ground truth.
Using a pessimistic hyphenation detection method (all potential hyphenation candidates), we shrink the training set to 80\,494 samples, which is just 73.42\% of the original one.
For the deviation prediction method, we keep the lines from the interval: $\left[\mu - \sigma, \mu + \sigma\right]$. We used one standard deviation as a trade-off between the quality and quantity of the leftover data.

Additionally, we consider two settings: Using the best model based on the CER on the validation set after 100 epochs (Difference-Full) and taking the model after training for only 10 epochs (Difference-10E) to investigate whether the model adapts too strongly to the existing error patterns in the ground truth. 
When we use Difference-Full, we are left with 89\,285 samples, which is 81.44\% of the original test set, and for Difference-10E with 81.76\%, respectively. 

\begin{table*}[!ht]
    \centering
    \begin{tabular}{ll|c|cc}
        \toprule
        {~} & {~} & Entire Test Set & \multicolumn{2}{c}{Corrected Test Set}  \\
        Training & Model                    & \textit{Original Labels} & \textit{Original Labels} & \textit{Corrected Labels} \\
        \midrule
        \multirow{2}{*}{Entire Training Set} & PyLaia & 9.13 & 19.47 & 12.89 \\
        ~ & TrOCR & 10.74 & 18.56 & 14.54 \\
        \hline
        \multirow{2}{*}{Exclude Hyphenations} & PyLaia & 9.71 & 18.39 & 13.12 \\
        ~ & TrOCR & 11.48 & 19.42 & 15.17 \\
        \hline
        \multirow{2}{*}{Exclude Difference-Full} & PyLaia & 9.61 & 18.59 & 13.40 \\
         ~ & TrOCR & 11.59 & 19.62 & 15.24 \\
         \hline
        \multirow{2}{*}{Exclude Difference-10E} & PyLaia & 9.41 & 18.21 & 13.03 \\ 
         ~ & TrOCR & 11.04 & 19.11 & 14.32 \\
        \bottomrule
    \end{tabular}
    \vspace{1em}
    \caption{\textbf{HTR results}. Character Error Rate (CER) for PyLaia and TrOCR when trained with different parts of the training set. The CER is indicated both for the entire test set as well as for the manually corrected part of the test set.}
    \label{tab:results}
\end{table*}

\newpage
\subsection{Corrected Test Set}
Out of the 1\,900 annotated text lines, only 415 (21.84\%) did have any errors that are a direct consequence of the hyphenation problem, but almost all of them could be fixed by the annotators as shown in \autoref{tab:annotation-stats-status}. In our experiments, we only use the 1\,878 lines that could be fixed, excluding 22 lines for which the annotators were unsure or unable to fix the error.

\begin{table}
\centering
\begin{tabular}{l|rr}
  \toprule
  \multicolumn{3}{c}{\textbf{Annotation Status}} \\
  \midrule
  Correct & 415 & (21.84\%) \\
  Fixed & 1\,463 & (77.00\%) \\
  Unsure & 21 & (1.11\%) \\
  Has Errors & 1 & (0.05\%) \\
  \hline
  \textbf{Total} & \textbf{1\,900} & \textbf{(100\%)} \\
  \bottomrule
\end{tabular}
\caption{Details of annotation status of the corrected test set.}\label{tab:annotation-stats-status}
\end{table}

At the start of the line, the most common errors are, that a word or part of it is missing from the ground truth, whereas at the end of the line, the most common case is the hyphenated words completed erroneously by a fairly big margin, even though other errors occur as well. Interestingly, there are 480 text lines, that contain an actual hyphenation character at the end of the line in the image, which is a quarter (25.26\%) of all annotated images as presented in \autoref{tab:annotation-stats-hyphen}.

\begin{table}
\centering
\begin{tabular}{l|rr|rr}
  \toprule
  \textbf{Labels} & \multicolumn{2}{c}{\textbf{Start of Line}} & \multicolumn{2}{|c}{\textbf{End of Line}} \\
  \midrule
  Missing Word(s) & 191 & (10.05\%) & 104 & (5.47\%) \\
  Additional Word(s) & 3  & (0.16\%) & 24 & (1.26\%) \\
  Hyphenated (missing) & 315 & (16.58\%) & 111 & (5.84\%) \\
  Hyphenated (extra chars) & 167 & (8.79\%) & 623 & (32.79\%) \\
  Hyphenation Character & \textendash{} & \textendash{} & 480 & (25.26\%) \\
  \bottomrule
\end{tabular}
\vspace{0.5em}
\caption{Types of errors that were corrected compared to the original ground truth.}\label{tab:annotation-stats-hyphen}
\end{table}

\subsection{HTR Results}
To investigate the impact of the errors on HTR, several experiments were conducted with PyLaia and TrOCR using 4 different training sets: The original entire training set, excluding hyphenated lines according to the proposed detection methods, and excluding text lines with a large character difference, both with a fully trained and a weakly trained HTR system. The evaluation is performed on 3 different test sets: The original entire test set, as well as the corrected part of the test set both with original and corrected labels, respectively.

The results are presented in \autoref{tab:results}. Focusing on the first row of the table (Entire Training Set), we can make the following observations:
\begin{itemize}
\item The corrected part of the test set, where we have explicitly focused on potential hyphenation problems, shows a dramatic loss in performance with respect to the original labels for both PyLaia and TrOCR. This indicates that our detection methods have allowed us to identify the systematic alignment errors we were looking for.
\item After correcting the labels, both the performance evaluations of PyLaia and TrOCR improve on the corrected part of the test set.
This indicates that the recognition outputs on the problematic lines were in fact better than one would have expected with the erroneous labels. Although we cannot be certain, because we cannot correct the entire test set, we assume that the HTR performance of 9.13\% for PyLaia and 10.74\% for TrOCR are actually underestimated and would improve for higher-quality test labels.
\end{itemize}

Focusing on the remaining rows of the table, we can observe the following:
\begin{itemize}
\item For PyLaia, the attempt to select fewer but higher-quality learning samples did not lead to an improved performance. In this case, keeping a larger quantity of learning samples was more successful, with the best result achieved for the entire training set (12.89\% CER).
\item For TrOCR, one of the proposed selection methods was able to slightly outperform the original training set: The difference-based selection using a weakly trained HTR system (14.32\% CER). It outperforms the fully trained HTR system, which might already be too fine-tuned to the systematic errors introduced by the alignment process.
\end{itemize}

\section{Conclusion}
\label{sec:conclusion}

Our experiments indicate a significant impact of alignment errors on the training and evaluation of deep-learning-based HTR systems. Automatic error detection and manual correction allowed us to gain a more realistic interpretation of the HTR results with respect to specific types of errors, especially wrongly hyphenated words.

Another observation relates to the trade-off between data quantity vs. quality in the training data, where the data quantity turned out to be more relevant. We have attempted to filter the training set, focusing on text lines with fewer errors. However, the loss of training data outweighed the benefit of high-quality learning samples. When compared to same-size training sets, a positive effect was visible but when compared with the entire training set, the HTR performance was significantly lower.

In future work, we aim to integrate more handwriting datasets with different types of systematic errors in a larger study. We aim to develop methods for detecting and correcting such errors while providing efficient tools for human users to efficiently perform corrections across large handwriting datasets. Also, we want to explore more modern handwritten and printed documents with different languages, as we think that algorithmic correction of ground truth seems very promising with more standardized languages. Finally, it would be interesting to complement training sets with synthetic learning samples~\cite{spoto2022improving}, where systematic errors can be avoided.

\subsection*{Acknowledgements}

This work has been supported by the Hasler Foundation, Switzerland.

\bibliographystyle{ACM-Reference-Format}
\bibliography{references}

\end{document}